\documentclass{article} 
\usepackage{iclr2016_conference,times}
\usepackage{hyperref}
\usepackage{url}
\usepackage{latexsym}
\usepackage{amsmath}
\usepackage{verbatim}
\usepackage{amssymb}
\usepackage{amsthm}
\usepackage{color}
\usepackage[pdftex]{graphicx}
\usepackage{multirow}
\usepackage{booktabs}
\iclrfinalcopy

\newcommand{\ignore}[1]{}

\title{Segmental Recurrent Neural Networks}

\author{
Lingpeng Kong, Chris Dyer \\
School of Computer Science\\
Carnegie Mellon University\\
Pittsburgh, PA 15213, USA \\
\texttt{\{lingpenk, cdyer\}@cs.cmu.edu}
\And
Noah A. Smith \\
Computer Science \& Engineering\\
University of Washington\\
Seattle, WA 98195, USA\\
\texttt{nasmith@cs.washington.edu}
}

\begin{document}

\maketitle

\begin{abstract}
We introduce \textbf{segmental recurrent neural networks} (SRNNs) which define, given an input sequence, a joint probability distribution over segmentations of the input and labelings of the segments. Representations of the input segments (i.e., contiguous subsequences of the input) are computed by encoding their constituent tokens using bidirectional recurrent neural nets, and these ``segment embeddings'' are used to define compatibility scores with output labels.
These local compatibility scores are integrated using a global semi-Markov conditional random field. Both fully supervised training---in which segment boundaries and labels are observed---as well as partially supervised training---in which segment boundaries are latent---are straightforward. Experiments on handwriting recognition and joint Chinese word segmentation/POS tagging show that, compared to models that do not explicitly represent segments such as BIO tagging schemes and connectionist temporal classification (CTC), SRNNs obtain substantially higher accuracies.\end{abstract}

\section{Introduction}
For sequential data like speech, handwriting, and DNA, segmentation and segment-labeling are abstractions that capture many common data analysis challenges.  We consider the joint task of breaking an input sequence into contiguous, arbitrary-length segments while labeling each segment.

Our new approach to this problem 
is the segmental recursive neural network (SRNN).  SRNNs combine two powerful machine learning tools:  representation learning and structured prediction.  First, bidirectional recurrent neural networks (RNNs) embed every feasible segment of the input in a continuous space, and these embeddings are then used to calculate the compatibility of each candidate segment with a label.
Unlike past RNN-based approaches (e.g., connectionist temporal classification or CTC; \citealp{graves:2006}) each candidate segment is represented explicitly, allowing application in settings where an alignment between segments and labels is desired as part of the output (e.g., protein secondary structure prediction or information extraction from text). 

At the same time, SRNNs are a variant of semi-Markov conditional random fields \citep{sarawagi:2004}, in that they define a conditional probability distribution over the output space (segmentation and labeling) given the input sequence (\S\ref{sec:model}). 
This allows explicit modeling of statistical dependencies, such as those between adjacent labels, and also of segment lengths (unlike widely used symbolic approaches based on ``BIO'' tagging; \citealp{ramshaw-marcus}).  Because the probability score decomposes into chain-structured clique potentials, polynomial-time dynamic programming algorithms exist for prediction and parameter estimation (\S\ref{sec:inference}).

Parameters can be learned with either a fully supervised objective---where both segment boundaries and segment labels are provided at training time---and partially supervised training objectives---where segment boundaries are latent (\S\ref{sec:learning}). 

We compare SRNNs to strong models that do not explicitly represent segments on handwriting recognition and joint word segmentation and part-of-speech tagging for Chinese text, showing significant accuracy improvements in both, demonstrating the value of models that explicitly model segmentation even when segmentation is not necessary for downstream tasks (\S\ref{sec:experiments}).

\section{Model}
\label{sec:model}
Given a sequence of input observations $\boldsymbol{x} = \langle \mathbf{x}_1, \mathbf{x}_2 , \ldots , \mathbf{x}_{|\boldsymbol{x}|} \rangle$ with length $|\boldsymbol{x}|$, a \textbf{segmental recurrent neural network} (SRNN) defines a joint distribution $p(\boldsymbol{y},\boldsymbol{z} \mid \boldsymbol{x})$ over a sequence of labeled segments each of which is characterized by a duration ($z_i \in \mathbb{Z}_+$) and label ($y_i \in Y$). The segment durations constrained such that $\sum_{i=1}^{|\boldsymbol{z}|} z_i = |\boldsymbol{x}|$. The length of the output sequence $|\boldsymbol{y}| = |\boldsymbol{z}|$ is a random variable, and $|\boldsymbol{y}| \le |\boldsymbol{x}|$ with probability 1. We write the starting time of segment $i$ as $s_i = 1+\sum_{j < i} z_j$.

To motivate our model form, we state several desiderata. First, we are interested in the following prediction problem,
\begin{align}
\boldsymbol{y}^* &= \arg \max_{\boldsymbol{y}} p(\boldsymbol{y} \mid \boldsymbol{x}) 
= \arg \max_{\boldsymbol{y}}  \sum_{\boldsymbol{z}}   p(\boldsymbol{y},\boldsymbol{z} \mid \boldsymbol{x}) 
\approx \arg \max_{\boldsymbol{y}}  \max_{\boldsymbol{z}}   p(\boldsymbol{y},\boldsymbol{z} \mid \boldsymbol{x}). \label{eq:max}
\end{align}
Note the use of joint maximization over $\boldsymbol{y}$ and $\boldsymbol{z}$ as a computationally tractable substitute for marginalizing out $\boldsymbol{z}$; this is commonly done in natural language processing.

\begin{figure}[h]
\centering
\includegraphics[scale=0.66]{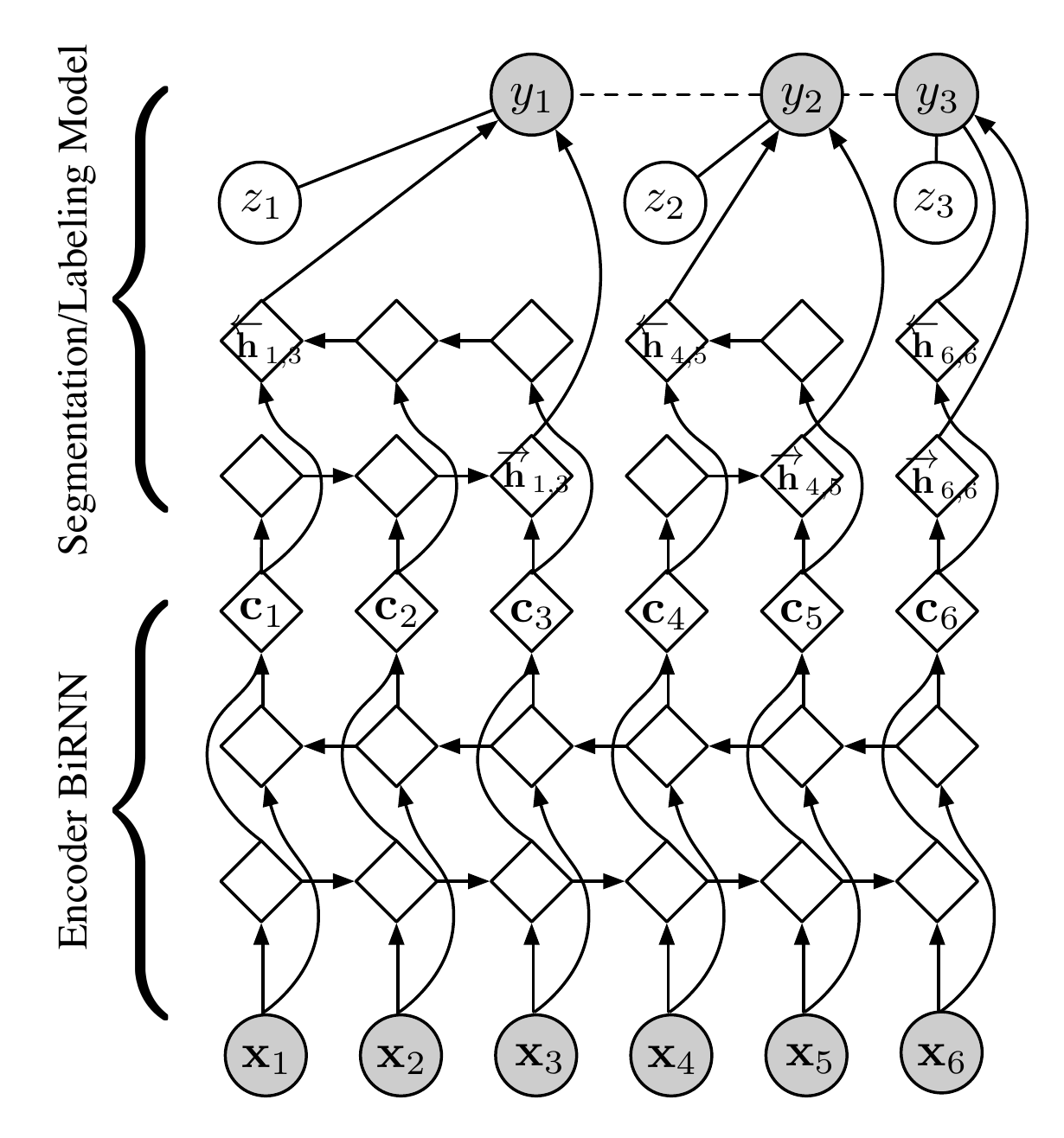}
\caption{Graphical model showing a six-frame input and three output segments with durations $\boldsymbol{z}=\langle 3,2,1 \rangle$ (this particular setting of $\boldsymbol{z}$ is shown only to simplify the layout of this figure; the model assigns probabilities to all valid settings of $\boldsymbol{z}$). Circles represent random variables. Shaded nodes are observed in training; open nodes are latent random variables; diamonds are deterministic functions of their parents; dashed lines indicate optional statistical dependencies that can be included at the cost of increased inference complexity. The graphical notation we use here draws on conventions used to illustrate neural networks and graphical models.
\label{fig:undirected}}
\end{figure}

Second, for problems where the explicit durations observations are unavailable at training time and are inferred as a latent variable, we must be able to use a marginal likelihood training criterion,
\begin{align}
\mathcal{L} & = - \log p(\boldsymbol{y} \mid \boldsymbol{x}) = - \log \sum_{\boldsymbol{z}} p(\boldsymbol{y},\boldsymbol{z} \mid \boldsymbol{x}) .\label{eq:opt}
\end{align}
In Eqs.~\ref{eq:max}~and~\ref{eq:opt}, the conditional probability of the labeled segment sequence is (assuming $k$th order dependencies on $\boldsymbol{y}$):
\begin{align}
p(\boldsymbol{y},\boldsymbol{z} \mid \boldsymbol{x}) = \frac{1}{Z(\boldsymbol{x})}\prod_{i=1}^{|\boldsymbol{y}|} \exp f(y_{i-k:i},z_i, \mathbf{x})
\label{eq:factor}
\end{align}
where $Z(\boldsymbol{x})$ is an appropriate normalization function. To ensure the expressiveness of $f$ and the computational efficiency of the maximization and marginalization problems in Eqs.~\ref{eq:max}~and~\ref{eq:opt}, we use the following definition of $f$,
\begin{equation}
\begin{aligned}
f(y_{i-k:i},z_i, \mathbf{x}_{s_i:s_i+z_i - 1}) = \mathbf{w}^{\top}\phi(\mathbf{V} [\mathbf{g}_y&(y_{i-k}); \ldots ;\mathbf{g}_y(y_i);\mathbf{g}_z(z_i);\\
&\overrightarrow{\textrm{RNN}}(\mathbf{c}_{s_i:s_i+z_i - 1});\overleftarrow{\textrm{RNN}}(\mathbf{c}_{s_i:s_i+z_i - 1})] + \mathbf{a}) + b
\end{aligned}
\label{eq:neural}
\end{equation}
where $\overrightarrow{\textrm{RNN}}(\mathbf{c}_{s_i:s_i+z_i - 1})$ is a recurrent neural network that computes the \textbf{forward segment embedding} by ``encoding'' the $z_i$-length subsequence of $\boldsymbol{x}$ starting at index $s_i$,\footnote{Rather than directly reading the $\mathbf{x}_i$'s, each token is represented as the concatenation, $\mathbf{c}_i$, of a forward and backward over the sequence of raw inputs. This permits tokens to be sensitive to the contexts they occur in, and this is standardly used with neural net sequence labeling models~\citep{graves:2006}.} and $\overleftarrow{\textrm{RNN}}$ computes the reverse segment embedding (i.e., traversing the sequence in reverse order), and $\mathbf{g}_y$ and $\mathbf{g}_z$ are functions which map the label candidate $\boldsymbol{y}$ and segmentation duration $\boldsymbol{z}$ into a vector representation. The notation $[\mathbf{a};\mathbf{b};\mathbf{c}]$ denotes vector concatenation. Finally, the concatenated segment duration, label candidates and segment embedding are passed through a affine transformation layer parameterized by $\mathbf{V}$ and $\mathbf{a}$ and a nonlinear activation function $\phi$ (e.g., $\tanh$), and a dot product with a vector $\mathbf{w}$ and addition by scalar $b$ computes the log potential for the clique. Our proposed model is equivalent to a semi-Markov conditional random field with local features computed using neural networks. Figure~\ref{fig:undirected} shows the model graphically.

We chose bidirectional LSTMs~\citep{graves2005framewise} as the implementation of the RNNs in Eq.~\ref{eq:neural}. LSTMs~\citep{hochreiter1997long} are a popular variant of RNNs which have been seen successful in many representation learning problems~\citep{graves:2014,karpathy:2015}. Bidirectional LSTMs enable effective computation for embedings in both directions and are known to be good at preserving long distance dependencies, and hence are well-suited for our task.

\section{Inference with Dynamic Programming}
\label{sec:inference}
We are interested in three inference problems: (i) finding the most probable segmentation/labeling for a model given a sequence $\boldsymbol{x}$; (ii) evaluating the partition function $Z(\boldsymbol{x})$; and (iii) computing the posterior marginal $Z(\boldsymbol{x},\boldsymbol{y})$, which sums over all segmentations compatible with a reference sequence $\boldsymbol{y}$. These can all be solved using dynamic programming. For simplicity, we will assume zeroth order Markov dependencies between the $y_i$s. Extensions to the $k$th order Markov dependencies should be straightforward. \ignore{Other quantities of interest, such as the posterior over segmentation decisions, marginal probabilities of labeling decisions can likewise be computed with similar dynamic programs.} Since each of these algorithms relies on the forward and reverse segment embeddings, we first discuss how these can be computed before going on to the inference algorithms.

\subsection{Computing Segment Embeddings} 
Let the $\overrightarrow{\mathbf{h}}_{i,j}$ designate the $\overrightarrow{\mathrm{RNN}}$ encoding of the input span $(i,j)$, traversing from left to right, and let $\overleftarrow{\mathbf{h}}_{i,j}$ designate the reverse direction encoding using $\overleftarrow{\mathrm{RNN}}$. There are thus $O(|\boldsymbol{x}|^2)$ vectors that must be computed, each of length $O(|\boldsymbol{x}|)$. Naively this can be computed in time $O(|\boldsymbol{x}|^3)$, but the following dynamic program reduces this to $O(|\boldsymbol{x}|^2)$:
\begin{align*}
\overrightarrow{\mathbf{h}}_{i,i} & = \overrightarrow{\mathrm{RNN}}(\overrightarrow{\mathbf{h}}_0, \mathbf{c}_i) \\
 \overrightarrow{\mathbf{h}}_{i,j} & = \overrightarrow{\mathrm{RNN}}(\overrightarrow{\mathbf{h}}_{i,j-1}, \mathbf{c}_{j}) \\
\overleftarrow{\mathbf{h}}_{i,i} & = \overleftarrow{\mathrm{RNN}}(\overleftarrow{\mathbf{h}}_0, \mathbf{c}_i) \\
\overleftarrow{\mathbf{h}}_{i,j} & = \overleftarrow{\mathrm{RNN}}(\overleftarrow{\mathbf{h}}_{i+1,j}, \mathbf{c}_{i}) 
\end{align*}
The algorithm is executed by initializing in the values on the diagonal (representing segments of length 1) and then inductively filling out the rest of the matrix. In practice, we often can put a upper bound for the length of a eligible segment thus reducing the complexity of runtime to $O(|\boldsymbol{x}|)$. This savings can be substantial for very long sequences (e.g., those encountered in speech recognition). 

\subsection{Computing the most probable segmentation/labeling and $Z(\boldsymbol{x})$}
For the input sequence $\boldsymbol{x}$, there are $2^{|\boldsymbol{x}| - 1}$ possible segmentations and $O(|Y|^{|\boldsymbol{x}|})$ different labelings of these segments, making exhaustive computation entirely infeasible. Fortunately, the partition function $Z(\boldsymbol{x})$ may be computed in polynomial time with the following dynamic program:
\begin{align*}
\alpha_0 &= 1 \\
\alpha_{j} &= \sum_{i < j} \alpha_i \times  \\
&\qquad \ \ \sum_{y \in Y}\left( \exp\mathbf{w}^{\top}\phi(\mathbf{V} [g_y(y);g_z(z_i);
\overrightarrow{\textrm{RNN}}(\mathbf{c}_{s_i:s_i+z_i - 1});\overleftarrow{\textrm{RNN}}(\mathbf{c}_{s_i:s_i+z_i - 1})] + \mathbf{a}) + b\right).
\end{align*}
After computing these values, $Z(\boldsymbol{x}) = \alpha_{|\boldsymbol{x}|}$. By changing the summations to a $\max$ operators (and storing the corresponding $\arg \max$ values), the maximal \emph{a posteriori} segmentation/labeling can be computed.

Both the partition function evaluation and the search for the MAP outputs run in time $O(|\boldsymbol{x}|^2\cdot |Y|)$ with this dynamic program. Adding $n$th order Markov dependencies between the $y_i$s adds requires additional information in each state and increases the time and space requirements by a factor of $O(|Y|^n)$. However, this may be tractable for small $|Y|$ and $n$.

\paragraph{Avoiding overflow.} Since this dynamic program sums over exponentially many segmentations and labelings, the values in the $\alpha_i$ chart can become very large. Thus, to avoid issues with overflow, computations of the $\alpha_i$'s must be carried out in log space.\footnote{An alternative strategy for avoiding overflow in similar dynamic programs is to rescale the forward summations at each time step~\citep{rabiner:1989,graves:2006}. Unfortunately, in a semi-Markov architecture each term in $\alpha_i$ sums over different segmentations (e.g., the summation for $\alpha_2$ will have contain some terms that include $\alpha_1$ and some terms that include only $\alpha_0$), which means there are no common factors, making this strategy inapplicable.}

\subsection{Computing $Z(\boldsymbol{x},\boldsymbol{y})$} To compute the posterior marginal $Z(\boldsymbol{x},\boldsymbol{y})$, it is necessary to sum over all segmentations that are compatible with a label sequence $\boldsymbol{y}$ given an input sequence $\boldsymbol{x}$. To do so requires only a minor modification of the previous dynamic program to track how much of the reference label sequence $\boldsymbol{y}$ has been consumed. We introduce the variable $m$ as the index into $\boldsymbol{y}$ for this purpose. The modified recurrences are:
\begin{align*}
\gamma_0(0) &= 1 \\
\gamma_j(m) &= \sum_{i<j}  \gamma_{i}(m-1) \times  \\
&\qquad\qquad \left( \exp\mathbf{w}^{\top}\phi(\mathbf{V} [g_y(y_i);g_z(z_i); \overrightarrow{\textrm{RNN}}(\mathbf{c}_{s_i:s_i+z_i - 1});\overleftarrow{\textrm{RNN}}(\mathbf{c}_{s_i:s_i+z_i - 1})] + \mathbf{a}) + b \right).
\end{align*}
The value $Z(\boldsymbol{x},\boldsymbol{y})$ is $\gamma_{|\boldsymbol{x}|}(|\boldsymbol{y}|)$.\ignore{\footnote{We note that it is possible to efficiently compute the marginal quantities over a set of strings (rather than a single string) if this set can be compactly encoded in a finite-state automaton with states $Q$. This algorithm runs in time $O(|\boldsymbol{x}|^2 \cdot |Q|)$.}}


 
\section{Parameter Learning}
\label{sec:learning}
We consider two different learning objectives.

\subsection{Supervised learning}
\label{sec:fully}
In the supervised case, both the segment durations ($\boldsymbol{z}$) and their labels  ($\boldsymbol{y}$) are observed.
\begin{align*}
\mathcal{L} &= \sum_{(\boldsymbol{x},\boldsymbol{y},\boldsymbol{z}) \in \mathcal{D}} -\log p(\boldsymbol{y},\boldsymbol{z} \mid \boldsymbol{x}) \\
&= \sum_{(\boldsymbol{x},\boldsymbol{y},\boldsymbol{z}) \in \mathcal{D}} \log Z(\boldsymbol{x}) - \log Z(\boldsymbol{x},\boldsymbol{y},\boldsymbol{z})
\end{align*}
In this expression, the unnormalized conditional probability of the reference segmentation/labeling, given the input $\boldsymbol{x}$ is written as $Z(\boldsymbol{x},\boldsymbol{y},\boldsymbol{z})$.

\subsection{Partially supervised learning}
\label{sec:partially}
In the partially supervised case, only the labels are observed and the segments (the $\boldsymbol{z}$) are unobserved and marginalized.
\begin{align*}
\mathcal{L} &= \sum_{(\boldsymbol{x},\boldsymbol{y}) \in \mathcal{D}} -\log p(\boldsymbol{y} \mid \boldsymbol{x}) \\
&=  \sum_{(\boldsymbol{x},\boldsymbol{y}) \in \mathcal{D}} \sum_{\boldsymbol{z}\in \mathcal{Z}(\boldsymbol{x},\boldsymbol{y})} -\log p(\boldsymbol{y}, \boldsymbol{z} \mid \boldsymbol{x}) \\
&= \sum_{(\boldsymbol{x},\boldsymbol{y}) \in \mathcal{D}} \log Z(\boldsymbol{x}) - \log Z(\boldsymbol{x},\boldsymbol{y})
\end{align*}
For both the fully and partially supervised scenarios, the necessary derivatives can be computed using automatic differentiation or (equivalently) with backward variants of the above dynamic programs~\citep{sarawagi:2004}.

\section{Experiments}
\label{sec:experiments}
We present two sets of experiments to compare segmental recurrent neural networks
against models  that do not include explicit representations of segmentation.
For the handwriting recognition task, we consider connectionist temporal classification (CTC)~\citep{graves:2006}; for Chinese word segmentation, 
we consider BIO tagging.
In these experiments, we do not include Markovian dependencies between adjacent labels for our models or the baselines. 

\subsection{Online Handwriting Recognition}
\textbf{Dataset} We use the handwriting dataset from \cite{kassel1995comparison}. This dataset is an online collection of hand-written words from 150 writers. It is recorded as the coordinates $(x,y)$ at time $t$ plus special pen-down/pen-up notations.  We break the coordinates into strokes using the pen-down and pen-up notations. One character typically consists one or more contiguous strokes.\footnote{There are infrequent cases where one stroke can go across multiple characters or the strokes which form the character can be not contiguous. We leave those cases for future work.}


The dataset is split into train, development and test set following \cite{kassel1995comparison}. Table~\ref{tab:stakassel} presents the statistics for the dataset.

A well-know variant of this dataset was introduced by \citet{taskar2004max}. \citet{taskar2004max} selected a ``clean'' subset of about 6,100 words and rasterized and normalized the images of each letter. Then, the uppercased letters (since they are usually the first character in a word) are removed and only the lowercase letters are used. The main difference between our dataset and theirs is that their dataset is ``offline'' --- \citet{taskar2004max} mapped each character into a bitmap and treated the segmentation of characters as a preprocessing step.  We use the richer representation of the sequence of strokes as input.


\begin{table}[]
\centering

\begin{tabular}{c|cc}
\toprule
      & \multicolumn{1}{c}{\#words} & \#characters \\ \midrule
Train & 4,368                        & 37,247       \\ 
Dev   & 1,269                        & 10,905       \\ 
Test  & 637                          & 5,516        \\ \midrule
Total & 6,274                        & 53,668       \\ \bottomrule
\end{tabular}
\caption{Statistics of the Online Handwriting Recognition Dataset\label{tab:stakassel}}
\end{table}

\textbf{Implementation}
We trained two versions of our model on this dataset, namely, the fully supervised model (\S \ref{sec:fully}), which takes advantage of the gold segmentations on training data,  and the partially supervised model (\S \ref{sec:partially}) in which the gold segmentations are only used in the evaluation. A CTC model reimplemented on the top of our Encoder BiRNNs layer (Figure~\ref{fig:undirected}) is used as a baseline so that we can see the effect of explicitly representing the segmentation.\footnote{The CTC interpretation rules specify that repeated symbols, e.g. {\tt aa} will be interpreted as a single token of {\tt a}. However since the segments in the handwriting recognition problem are extremely short, we use different rules and interpret this as {\tt aa}. That is, only the blank symbol may be used to represent extended durations. Our experiments indicate this has little effect, and Graves (p.c.) reports that this change does not harm performance in general.} For the decoding of the CTC model, we simply use the best path decoding, where we assume that the most probable path will correspond to the most probable labeling, although it is known that prefix search decoding can slightly improve the results~\citep{graves:2006}.

As a preprocessing step, we first represented each point in the dataset using a 4 dimensional vector, $\mathbf{p} = (p_x, p_y, \Delta p_x, \Delta p_y)$, where $p_x$ and $p_y$ are the normalized coordinates of the point and $\Delta p_x$ and $\Delta p_y$ are the corresponding changes in the coordinates with respect to the previous point. $\Delta p_x$ and $\Delta p_y$ are meant to capture basic direction information. Then we map the points inside one stroke into a fixed-length vector using a bi-direction LSTM. Specifically, we concatenated the last position's hidden states in both directions and use it as the input vector $\mathbf{x}$ for the stroke.


In all the experiments, we use Adam~\citep{kingma2014adam} with $\lambda = 1 \times 10^{-6}$ to optimize the parameters in the models. We train these models until convergence and picked the best model over the iterations based on development set performance then report performance on the test set.

We used 5 as the hidden state dimension in the bidirectional RNNs, which map the points into fixed-length stroke embeddings (hence the input vector size $5 \times 2 = 10$). We set the hidden dimensions of $\mathbf{c}$ in our model and CTC model to 24 and segment embedding $\mathbf{h}$ in our model as 18. These dimensions were chosen based on intuitively reasonable values, and it was confirmed on development data that they performed well. We tried to experiment with larger hidden dimensions and we found the performance did not vary much. Future work might more carefully optimize these parameters.

As for speed, the partially supervised SRNNs run at $\sim$40 instances per second and the fully supervised SRNNs run at $\sim$53 instances during training using a single CPU.

\textbf{Results}
The results of the online handwriting recognition task are presented in Table~\ref{tab:hwr}. We see that both of our models outperform the baseline CTC model, which does not carry an explicit representation for the segments being labeled, by a significant margin. An interesting finding is, although the partially supervised model performs slightly worse in the development set, it actually outperforms the fully supervised model in the test set. 
Because the test set is written by different people from the train and development set, they exhibit different styles in their handwriting; our results suggest that
the partially supervised model may generalize better across different writing styles. 

\begin{table}[]
\centering
\begin{tabular}{c|cccc|cccc}
\toprule
\multirow{2}{*}{} & \multicolumn{4}{c|}{Dev}              & \multicolumn{4}{c}{Test}            \\
                  & $P_{seg}$  & $R_{seg}$  & $F_{seg}$  & Error      & $P_{seg}$  & $R_{seg}$  & $F_{seg}$  & Error     \\ \midrule
SRNNs (Partial)   & 98.7\% & 98.4\% & 98.6\% & 4.2\%  & 99.2\% & 99.1\% & 99.2\% & 2.7\% \\ 
SRNNs (Full)      & 98.9\% & 98.6\% & 98.8\% & 4.3\%  & 98.8\% & 98.6\% & 98.6\% & 5.4\% \\ 
CTC               & -       & -       & -       & 15.2\% & -       & -       & -       & 13.8\% \\
\bottomrule

\end{tabular}
\caption{Hand-writing Recognition Task\label{tab:hwr}}
\end{table}


\subsection{Joint Chinese Word Segmentation and POS tagging}
In this section, we will look into two related tasks. The first task is joint Chinese word segmentation and POS tagging, where the $\boldsymbol{z}$ variables will group the Chinese characters into words and the $\boldsymbol{y}$ variables assign POS tags as labels to these words. We also tested our model on pure Chinese word segmentation task, where the assignments of $\boldsymbol{z}$ is the only thing we care about (simulated using a single label for all segments). 

\textbf{Dataset} We used standard benchmark datasets for these two tasks. For the joint Chinese word segmentation and POS tagging task, we use the Penn Chinese Treebank 5~\citep{xue2005penn}, following the standard train/dev/test splits. For the pure Chinese word segmentation task, we used the SIGHAN 2005 dataset\footnote{\url{http://www.sighan.org/bakeoff2005/}}. This dataset contains four portions, covering both simplified and traditional Chinese. Since there is no pre-assigned dev set in this dataset (only train and test set are provided), we manually split the original train set into two, one of which (roughly the same size as the test set) is used as the dev set. For both tasks, we use Wang2Vec~\citep{ling2015two} to generate the pre-trained character embeddings from the Chinese Gigaword~\citep{graff2005chinese}.

\textbf{Implementation}
Only supervised version SRNNs (\S \ref{sec:fully}) is tested in these tasks. The baseline model is a bi-directional LSTM tagger (basically the same structure as our Encoder BiRNNs in Figure~\ref{fig:undirected}). It takes the $\mathbf{c}$ at each time step and pushes it through an element-wise non-linear transformation ($\tanh$) followed by an affine transformation to map it to the same dimension as the number of labels. The total loss is therefore the sum of negative log probabilities over the sequence. Greedy decoding is applied in the baseline model, making it a zeroth order model like our SRNNs.

In order to perform segmentation and POS tagging jointly, we composed the POS tags with ``B'' or ``I'' to represent the segmentation point. For the segmentation-only task, in the SRNNs we simply used same dummy tag for all $y$ and only care about the $z$ assignments. In the BiRNN case, we used ``B'' and ``I'' tags.

For both tasks, the dimension for the input character embedding is 64. For our model, the dimension for $\mathbf{c}$ and the segment embedding $\mathbf{h}$ is set to 24. For the baseline bi-directional LSTM tagger, we set the hidden dimension (the $\mathbf{c}$ equivalent) size to 128. Here we deliberately chose a larger size than in our model hoping to make the number of parameters in the bi-directional LSTM tagger roughly the same as our model. We trained these models until convergence and picked the best model over iterations based on its performance on the development set.

As for speed, the SRNNs run at $\sim$3.7 sentence per second during training on the CTB dataset using a single CPU.

\textbf{Results} 
Table~\ref{tab:joint} presents the results for the joint Chinese word segmentation task. We can see that in both segmentation and POS tagging, the SRNNs achieve higher $F$-scores than the BiRNNs.

Table~\ref{tab:cseg} presents the results for the pure Chinese word segmentation task. The SRNNs perform better than the BiRNNs with the exception of the PKU portion of the dataset. The reason for this is probably because the training set in this portion is the smallest among the four. Thus leads to high variance in the test results.

\begin{table}[]
\centering
\begin{tabular}{c|ccc|ccc}
\toprule
                                & \multicolumn{3}{c|}{Dev}     & \multicolumn{3}{c}{Test}   \\
                                & $P_{seg}$  & $R_{seg}$  & $F_{seg}$  & $P_{seg}$  & $R_{seg}$  & $F_{seg}$  \\ \midrule
\multicolumn{1}{c|}{BiRNNs} & 93.2\% & 92.9\% & 93.0\% & 94.7\% & 95.2\% & 95.0\% \\
\multicolumn{1}{c|}{SRNNs}  & 93.8\% & 93.8\% & 93.8\% & 95.3\% & 95.8\% & 95.5\% \\ \midrule
                                & $P_{tag}$  & $R_{tag}$  & $F_{tag}$  & $P_{tag}$  & $R_{tag}$  & $F_{tag}$  \\ \midrule
\multicolumn{1}{c|}{BiRNNs} & 87.1\% & 86.9\% & 87.0\% & 88.1\% & 88.5\% & 88.3\% \\
\multicolumn{1}{c|}{SRNNs}  & 89.0\% & 89.1\% & 89.0\% & 89.8\% & 90.3\% & 90.0\% \\ \bottomrule
\end{tabular}
\caption{Joint Chinese Word Segmentation and POS Tagging 
\label{tab:joint}}
\end{table}

\begin{table}[]
\centering
\begin{tabular}{c|ccc|ccc}
\toprule
    & \multicolumn{3}{c|}{BiRNNs}  & \multicolumn{3}{c}{SRNNs} \\ 
    & $P_{seg}$       & $R_{seg}$       & $F_{seg}$       & $P_{seg}$       & $R_{seg}$       & $F_{seg}$       \\ \midrule
CU  & 92.7\% & 93.1\% & 92.9\% & 93.3\% & 93.7\% & 93.5\%  \\ 
AS  & 92.8\% & 93.5\% & 93.1\% & 93.2\% & 94.2\% & 93.7\%  \\
MSR & 89.9\% & 90.1\% & 90.0\% & 90.9\% & 90.4\% & 90.7\%  \\
PKU & 91.5\% & 91.2\% & 91.3\% & 90.6\% & 90.6\% & 90.6\%  \\ \bottomrule
\end{tabular}
\caption{Chinese Word Segmentation Results on SIGHAN 2005 dataset. There are four portions of the dataset from City University of Hong Kong (CU), Academia Sinica (AS), Microsoft Research (MSR) and Peking University (PKU). The former two are in traditional Chinese and the latter two are in simplified Chinese. 
\label{tab:cseg}
}
\end{table}


\section{Related Work}
\label{sec:related}
Segmental labeling problems have been widely studied. A widely used approach to a segmental labeling problems with neural networks is the connectionist temporal classification (CTC) objective and decoding rule of \cite{graves:2006}.  CTC reduces the ``segmental'' sequence label problem to a classical sequence labeling problem in which every position in an input sequence $\boldsymbol{x}$ is explicitly labeled by interpreting repetitions of input labels---or input labels followed by a special ``blank'' output symbol---as being a single label with a longer duration. During training, the marginal likelihood of the set of labelings compatible (according to the CTC interpretation rules) with the reference label $\boldsymbol{y}$ is maximized. CTC has demonstrated impressive success in various fully discriminative end-to-end speech recognition models \citep[\emph{inter alia}]{graves:2014,maas:2015,hannun:2014}.

Although CTC has been used successfully and its reuse of conventional sequence labeling architectures is appealing, it has several potentially serious limitations. First, it is not possible to model interlabel dependencies explicitly---these must instead be captured indirectly by the underlying RNNs. Second, CTC has no explicit segmentation model. Although this is most serious in applications where segmentation is a necessary/desired output (e.g., information extraction, protein secondary structure prediction), we argue that explicit segmentation is potentially valuable even when the segmentation is \emph{not} required. To illustrate the value of explicit segments, consider the problem of phone recognition. For this task, segmental duration is strongly correlated with label identity (e.g., while an [o] phone token might last 300ms, it is unlikely that a [t] would) and thus modeling it explicitly may be useful. Finally, making an explicit labeling decision for every position (and introducing a special blank symbol) in an input sequence is conceptually unappealing.

Several alternatives to CTC have been approached, such as using various attention mechanisms in place of marginalization \citep{chan:2015,bahdanau:2015}. These have been applied to end-to-end discriminative speech recognition problem. A more direct alternative to our method---indeed it was proposed to solve several of the same problems we identified---is due to \cite{graves:2012}. However, a crucial difference is that our model explicitly constructs representations of segments which are used to label the segment while that model relies on a marginalized frame-level labeling with a null symbol. 

The work of~\cite{abdel-hamid:2013} also seeks to construct embeddings of multi-frame segments. Their approach is quite different than the one taken here. First, they compute representations of variable-sized segments by uniformly sampling a fixed number of frames and using these to construct a representation of the segment with a simple feedforward network. Second, they do not consider them problem of latent segmentation.

Finally, using neural networks to provide local features in conditional random field models has also been proposed for sequential models~\citep{peng:2009} and tree-structured models~\citep{Durrett-Klein:2015:NeuralCRF}. To our knowledge, this is the first application to semi-Markov structures.

\section{Conclusion}
\label{sec:conclusion}
We have proposed a new model for segment labeling problems that learns representations of segments of an input sequence and then labels these. We outperform existing alternatives both when segmental information should be recovered and when it is only latent. We have not trained the segmental representations to be of any use beyond making good labeling (or segmentation) decisions, but an intriguing avenue for future work would be to construct representations that are useful for other tasks.

\subsubsection*{Acknowledgments}
The authors thank the anonymous reviewers, Yanchuan Sim, and Hao Tang for their helpful feedback. This work was sponsored in part by the Defense Advanced Research Projects Agency (DARPA)
Information Innovation Office (I2O) under the Low Resource Languages for Emergent Incidents (LORELEI) program issued by DARPA/I2O under Contract No.~HR0011-15-C-0114.

\bibliography{biblio}
\bibliographystyle{iclr2016_conference}

\end{document}